\documentclass[10pt,twocolumn,letterpaper]{article}

\usepackage[pagenumbers]{cvpr} 
\usepackage[accsupp]{axessibility}  
\definecolor{cvprblue}{rgb}{0.21,0.49,0.74}
\usepackage[pagebackref,breaklinks,colorlinks,allcolors=cvprblue]{hyperref}

\def\paperID{57} 
\def\confName{CVPR}
\def\confYear{2025}

\title{Semantic Matters: Multimodal Features for Affective Analysis}

\author{Tobias Hallmen\\
Chair for Human-Centered Artificial Intelligence\\
University of Augsburg\\
{\tt\small tobias.hallmen@uni-a.de}
\and
Robin-Nico Kampa\\
Institute for Distributed Intelligent Systems\\
University of the Bundeswehr Munich\\
{\tt\small robin-nico.kampa@unibw.de}
\and
Fabian Deuser\\
Institute for Distributed Intelligent Systems\\
University of the Bundeswehr Munich\\
{\tt\small fabian.deuser@unibw.de}
\and
Norbert Oswald\\
Institute for Distributed Intelligent Systems\\
University of the Bundeswehr Munich\\
{\tt\small norbert.oswald@unibw.de}
\and
Elisabeth André\\
Chair for Human-Centered Artificial Intelligence\\
University of Augsburg\\
{\tt\small elisabeth.andre@uni-a.de}
}

\begin{document}
\maketitle




\begin{abstract}

In this study, we present our methodology for two tasks: the Emotional Mimicry Intensity (EMI) Estimation Challenge and the Behavioural Ambivalence/Hesitancy (BAH) Recognition Challenge, both conducted as part of the 8th Workshop and Competition on Affective \& Behavior Analysis in-the-wild. We utilize a Wav2Vec 2.0 model pre-trained on a large podcast dataset to extract various audio features, capturing both linguistic and paralinguistic information. Our approach incorporates a valence-arousal-dominance (VAD) module derived from Wav2Vec 2.0, a BERT text encoder, and a vision transformer (ViT) with predictions subsequently processed through a long short-term memory (LSTM) architecture or a convolution-like method for temporal modeling. We integrate the textual and visual modality into our analysis, recognizing that semantic content provides valuable contextual cues and underscoring that the meaning of speech often conveys more critical insights than its acoustic counterpart alone. Fusing in the vision modality helps in some cases to interpret the textual modality more precisely. This combined approach results in significant performance improvements, achieving in EMI $\rho_{\text{TEST}} = 0.706$ and in BAH $F1_{\text{TEST}} = 0.702$, securing first place in the EMI challenge and second place in the BAH challenge.
\end{abstract}

\section{Introduction}
Affective computing enables deeper insights into human behavior and the complex dynamics of human interaction. By modeling and interpreting affective signals, it supports advances in fields such as psychology and e-learning. Crucially, it enables scalable, continuous observation, making it possible to assess factors such as student engagement and comprehension in real time, even in scenarios where direct human supervision is not feasible.

Given the growing importance of affective computing in understanding human behavior, the Affective \& Behavior Analysis in-the-wild Workshop~\cite{kollias2019deep, kollias2020analysing, kollias2021affect, kollias2021analysing, kollias2022abaw, kollias2023abaw, kollias2023abaw2, zafeiriou2017aff, kollias2024distribution, kollias20246th, kollias20247th} has played a key role in advancing the field over recent years. It provides a platform for tackling tasks such as action unit detection, emotion recognition, compound emotion analysis, and valence-arousal prediction. In this work we provide approaches for two task of the 8th iteration~\cite{Kollias2025, kolliasadvancements} of the ABAW Workshop, namely Emotional Mimicry Intensity (EMI) estimation and Behavioural Ambivalence/Hesitancy (BAH) recognition.

Emotional mimicry refers to the tendency of individuals to imitate the emotional expressions of others within social interactions. This behavior is particularly accurate when the observed person is familiar or well-liked~\cite{hess2014emotional}. By analyzing and predicting emotional mimicry using deep learning methods, we can better interpret social contexts because these models help reveal the nature and quality of relationships between individuals. In this challenge the HUME-Vidmimic2 dataset~\cite{christ2023muse} is utilized which features more than 30 hours of audiovisual recordings featuring 557 participants. The dataset is partitioned into training (8072 videos, approximately 15 hours), validation (4588 videos, approximately 9 hours), and test subsets (4582 videos, approximately 9 hours), with each participant appearing in only one subset. In each video participants are advised to mimic emotions and later label the intensity of the felt emotions as annotation. Provided data include face detections at 6 frames per second, visual features derived from Vision Transformer (ViT) models~\cite{caron2021emerging, dosovitskiy2020image}, audio representations extracted from Wav2Vec 2.0~\cite{baevski2020wav2vec}, as well as the raw audio and video files. The dataset comprises annotations for six distinct emotional expressions: ``Admiration'', ``Amusement'', ``Determination'', ``Empathic Pain'', ``Excitement'', and ``Joy''. The performance on the task is quantified using the mean Pearson’s Correlation Coefficient $\rho$ across all emotions, calculated as follows:
\begin{equation*}
    \rho_{\text{VAL}} = \frac{1}{6}\sum_{i=1}^6{\rho_i}, \, \rho_i = \frac{Cov(X_{i, pred}, Y_{i, label})}{\sqrt{Var(X_{i, pred})}\sqrt{Var(Y_{i, label})}}
\end{equation*}

The second task addressed in this work is the detection of Behavioral Ambivalence/Hesitancy (BAH). BAH refers to the presence of conflicting emotions or intentions expressed by individuals, particularly in the context of behavior change~\cite{Kollias2025, kolliasadvancements}. Using deep learning, we aim to automatically detect subtle verbal and nonverbal indicators of ambivalence and hesitation in participants responses. By learning to recognize these complex patterns, such models can support more nuanced analysis in the development of targeted interventions. The provided audiovisual dataset, which consists of 646 videos of 124 subjects who answered up to seven predefined questions with a total duration of approximately $5.3$ hours and approximately 465K frames. This is then randomly split into 84 subjects ($3.4$ hours, 295k frames) in train, respectively 40 subjects ($1.9$ hours, 170k frames) in test. These are annotated for absence (0) and presence (1) of ambivalence or hesitancy on a frame level following a codebook including facial expressions, audio cues, language, and body language. Textual transcriptions are also provided to participants. Performance on this task is evaluated using the weighted frame-level F1 score:

\begin{equation*}
    F1 = \frac{\sum_{i=0}^1 |n_i| F1_i}{\sum_{i=0}^1 |n_i|}, \,
    F1_i = \frac{2\cdot \text{TP}_i}{2\cdot \text{TP}_i + \text{FP}_i + \text{FN}_i}
\end{equation*}

To develop a unified solution for both tasks, we draw upon prior work in action unit detection~\cite{kollias2019face}, multi-task learning for affect and expression recognition~\cite{kollias2023multi, kollias2019expression, kollias2024distribution}, as well as our recent work on modeling emotional mimicry intensity~\cite{hallmen2024unimodal}. In our previous study, we compared the audio and visual modalities and found that unimodal prediction using audio provided the strongest performance. In this work, we extend our approach by incorporating textual context through transcriptions of the audio data. Interestingly, we find that semantic information derived from text serves as an even stronger prior, outperforming both raw audio and visual inputs. This suggests that current audio models are limited in their ability to capture the deeper meaning, context, and structure conveyed in speech. Therefore, we analyze how different modalities contribute to prediction performance and use a multi-task assesment framework~\cite{hallmen2023phoneme} to better understand the role each modality plays in the learning process.
\begin{figure*}[t]
\begin{center}
     \includegraphics[width=1.0\linewidth]{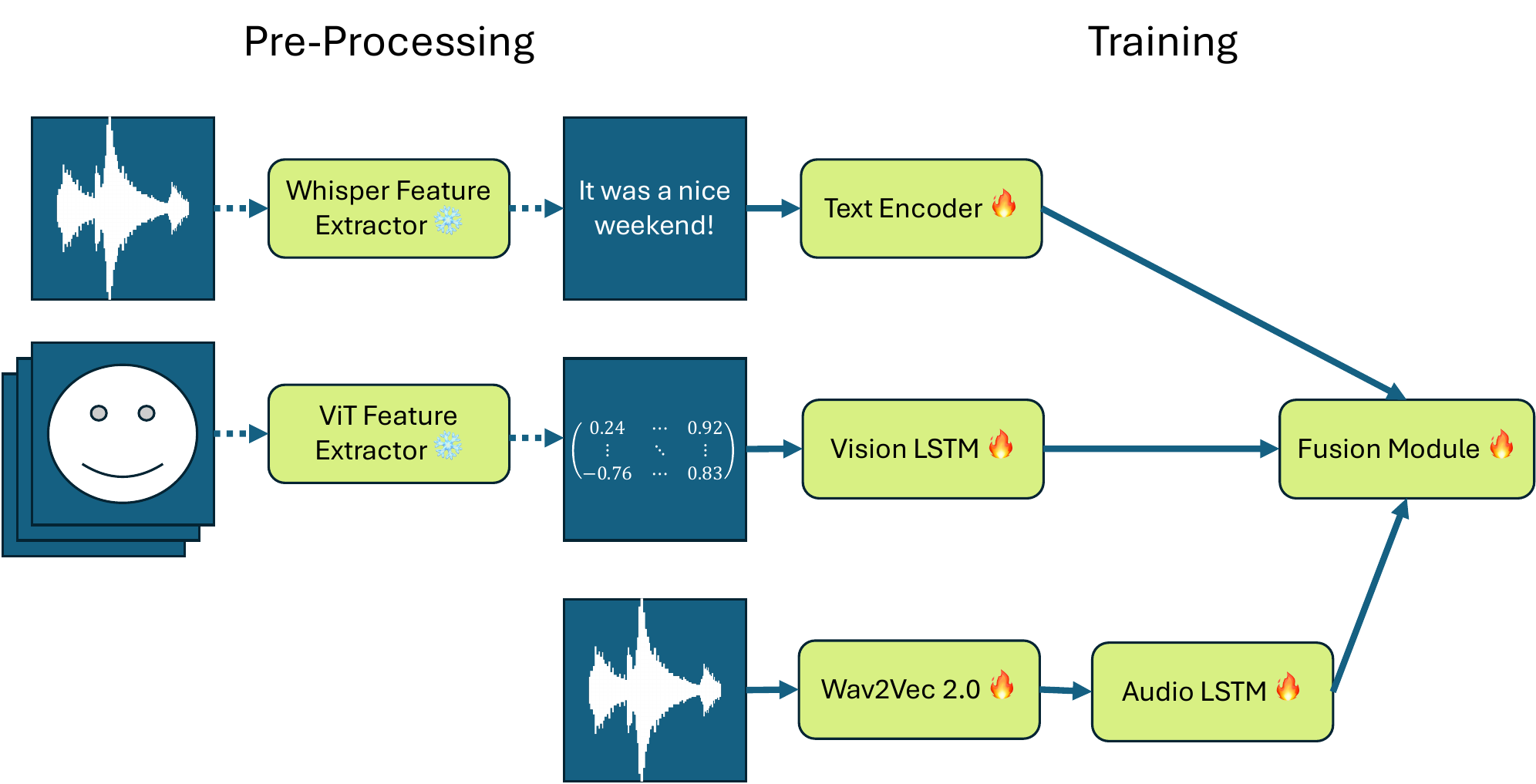}
    \end{center}
      \caption{\textbf{Architecture overview of our approach.} We first pre-process the cropped face images and transcribe the audio. Afterwards we process each modality independently and then fuse it in our fusion module.}
      \label{fig:method}
\end{figure*}
\section{Related Work}
Recent approaches to emotional mimicry intensity (EMI) prediction use multimodal embeddings that combine audio and visual cues. Savchenko et al.~\cite{savchenko2024hsemotion} aggregate statistics (mean, standard deviation, minimum, and maximum) from audio and facial embeddings extracted via Wav2Vec 2.0~\cite{baevski2020wav2vec} and pre-trained facial affective models, respectively. These features are concatenated and classified with a linear feed-forward layer using sigmoid activations. Yu et al.~\cite{yu2024efficient} use a ResNet-18 encoder~\cite{he2016deep} for facial features that predicts facial action units (AUs) while extracting audio embeddings from Wav2Vec 2.0. Temporal dependencies are captured by a temporal convolutional network (TCN), and modality fusion is performed by late fusion. Hallmen et al.~\cite{hallmen2024unimodal} rely solely on audio embeddings obtained from a Wav2Vec 2.0 model pre-trained specifically for emotional speech~\cite{wagner_2022_6221127}, using an LSTM for temporal modeling while incorporating additional features from Valence-Arousal-Dominance (VAD) predictions. Zhang et al.~\cite{zhang2024affective} propose a multimodal ensemble strategy, training a modality-specific Masked Autoencoder (MAE)~\cite{he2022masked} for facial representations and using several pre-trained audio encoders (VGGish~\cite{hershey2017cnn}, Wav2Vec 2.0, and HuBERT~\cite{hsu2021hubert}) for robust EMI prediction.

For the ambivalence and hesitancy prediction task, Richet et al.~\cite{richet2024textualized} propose a baseline utilizing textual features derived from speech transcriptions generated by a Whisper model~\cite{radford2023robust} and encoded with BERT-base~\cite{devlin2019bert}, audio embeddings from a pretrained Wav2Vec 2.0 model, and visual embeddings from facial crops encoded by a ResNet-50 pretrained on large-scale facial datasets~\cite{guo2016ms}. Each modality is individually processed using TCNs, and subsequently integrated via a co-attention mechanism~\cite{zhang2022continuous} to perform the final prediction.
In this year's iteration of the ABAW workshop, Savchenko et al.~\cite{savchenko2024hsemotion} use facial features from EmotiEffLib~\cite{savchenko2023facial} while incorporating the provided Wav2Vec 2.0 features. In addition, they show that statistical feature pooling outperforms simple average pooling when aggregating features from video frames.
Yu et al.~\cite{yu2025dual} present a two-stage architecture. The first stage uses a CLIP-like~\cite{radford2023robust} training framework to align multiple modalities. The second stage uses a Temporal Convolutional Network (TCN) in conjunction with a BiLSTM to effectively capture the temporal dynamics of video, audio, and text features.
\section{Methodology}
In our methodology, code publicly available here\footnote{\url{https://github.com/Skyy93/CVPR2025_abaw}},
we utilize all three available modalities, namely vision, text and audio as shown in Figure~\ref{fig:method}.

We encode the extracted face images from the videos using a ViT-Huge model~\cite{wu2020visual} pre-trained on ImageNet-21k. We do not perform any further fine-tuning on the ViT-Huge model, nor do we apply any additional augmentations to the input images. The ViT uses a patch size of 14, which allows it to capture more fine-grained visual features. These extracted feature sequences are then fed into a two-layer LSTM to model temporal dependencies. 

The final hidden state from the LSTM is then fed into our fusion module, which integrates modality-specific features (text, audio, and vision) using a two-layer MLP with a Tanh activation function, and ultimately predicts the logits.

For the audio features, we utilize a pre-trained Wav2Vec 2.0 model~\cite{wagner_2022_6221127}. An advantage of this model is its pre-training on the MSP podcast dataset~\cite{Lotfian_2019_3}, which contains natural emotional speech, thus enables the model to learn robust features tailored for affective computing tasks. In addition to extracting audio features, the model also produces Valence, Arousal and Dominance (VAD) predictions, which provide additional emotional context. Both the generated hidden states and the VAD predictions are then fed into a two-layer LSTM to model temporal dynamics before being integrated into the fusion module.

Finally, textual information is extracted from the videos using the Whisper large-v3 model~\cite{radford2023robust} and subsequently encoded with the GTE text encoder~\cite{zhang2024mgte}. In our experiments, adding an extra LSTM for temporal modeling, similar to our approach for visual and audio modalities, did not improve performance. Therefore, we directly use the pooled [CLS] token embedding as the textual representation and integrate it into the fusion module as additional features.
\subsection{EMI}

In the EMI challenge, we use 12-second audio segments as input for training because 95\% of the training data does not exceed this duration. Longer sequences did not improve performance. For the text modality, we extract 128 tokens from Whisper-generated transcripts, while for video we use a maximum of 400 frames, which also covers 95\% of the training data. Shorter sequences are padded to these respective maximums.

Our training setup employs a learning rate of 1e-4 with cosine decay. We optimize using Mean Squared Error (MSE) loss and train for 30 epochs with a batch size of 32. Early stopping on the validation set is used to determine the best checkpoint.
\begin{figure}[t]
\begin{center}  \includegraphics[width=\columnwidth]{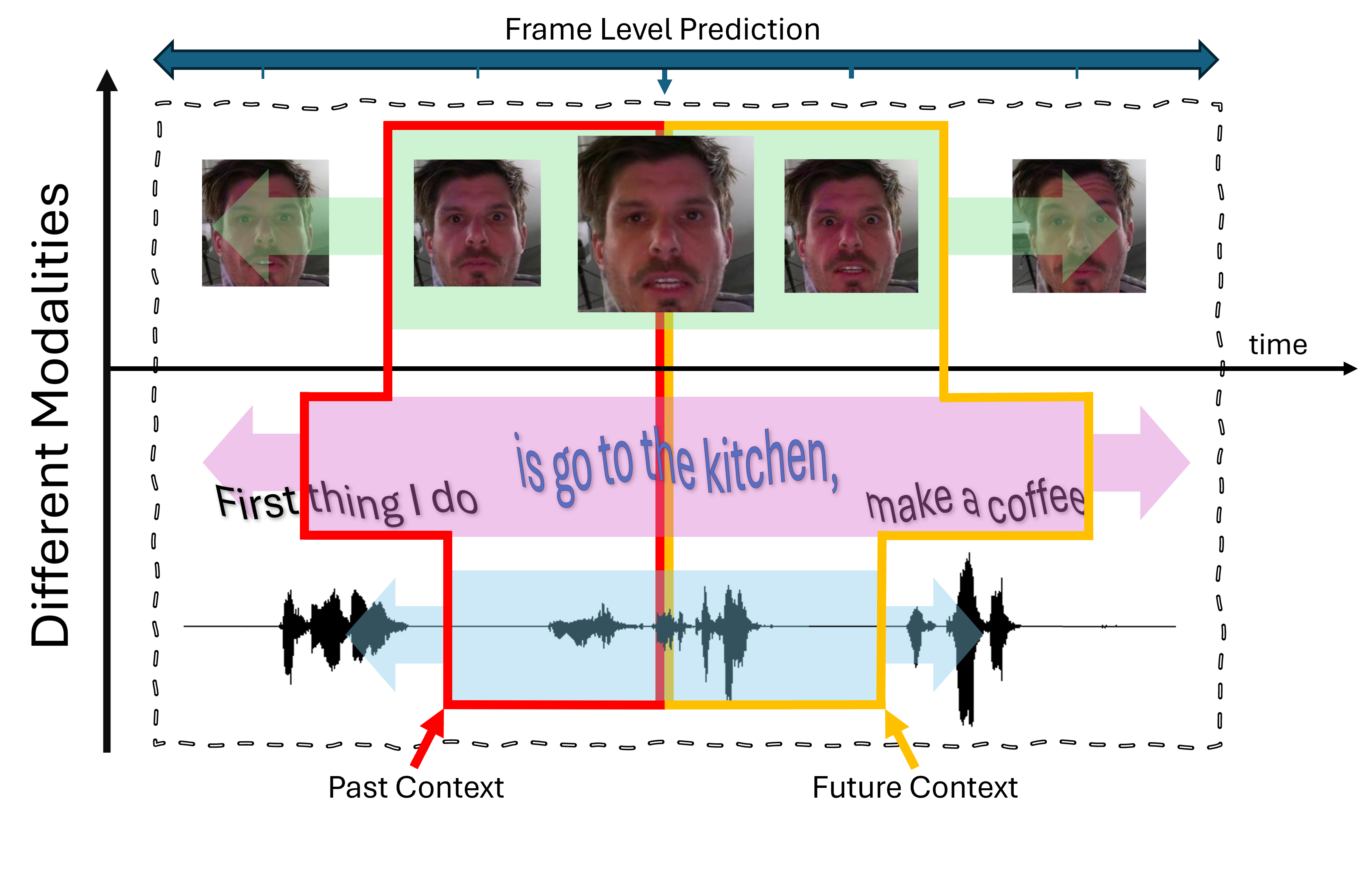}
    \end{center}
      \caption{\textbf{Using chunks from different modalities.} The chunk sizes are chosen differently for each modality, based on achieving optimal performance on the validation split. The available chunk sizes and their variations are depicted as colored boxes with arrows in the figure. The data shown are taken from the BAH challenge dataset.}
      \label{fig:bah-approach}
\end{figure}
\subsection{BAH}
Unlike traditional sequence-to-sequence modeling as in \cite{sutskever2014sequence}, our approach is more interpretable by operating as a deep learning-based convolution over multiple modalities. In other words, it enables context-aware sequence-to-point estimation at frame level. It integrates information from three modalities - audio, text, and images - across separately adjustable temporal windows that span both the past and the future. The data is preprocessed so that the current time point is centered within all three temporal windows, with the surrounding data providing context for the model to make predictions at each time step (e.g., per frame). When processing entire sequences, this method is very similar to classical convolution, as illustrated in Figure~\ref{fig:bah-approach}, which shows the contextual capture of different chunks at a specific time point. 

Apart from centering the data, we also introduced a special token to mark the current time point in the processed text, enabling the model to learn this special token during training. Apart from the convolution-like approach for our input data the models used are identical to our EMI approach shown in Figure~\ref{fig:method}. The temporal window for the text operates at word level.

For our training we use a learning rate of 7.5 $\cdot$ 1e-6 with cosine decay and employ an classical Binary Cross Entropy as loss. The model is trained for 10 epochs with a batch size of 32, and early stopping based on validation performance is employed to determine the optimal checkpoint. 

Attempts were also made to smooth the prediction results by applying filters. However, these efforts did not lead to any further improvements. In general, this aspect requires further investigation.

\section{Evaluation}

\subsection{EMI}
As shown in Table~\ref{tab:eval}, our proposed methods outperform both the baseline and two other approaches~\cite{savchenko2024hsemotion, yu2025dual} from this year's EMI challenge. For our predictions we use the best modals derived from our ablation study in section~\ref{sec:ablationemi}. Through extensive experiments with different modalities, we find that textual information provides the strongest features for the EMI task, despite being extracted from audio. Interestingly, although the audio model uses Wav2Vec 2.0 - pre-trained on natural emotional speech and designed for ASR tasks - it seems to have a relatively limited understanding of spoken content for the EMI context. In addition, combining audio with other modalities often reduced rather than improved performance. Visual features alone showed very limited effectiveness. However, when combined with text, vision achieved the highest correlation ($\rho$) on the validation set - apparently to pick up some of the nuances text is unable to, vision does help, despite being the weakest modality on its own. 
\begin{figure}[]
\begin{center}
     \includegraphics[width=1.0\columnwidth]{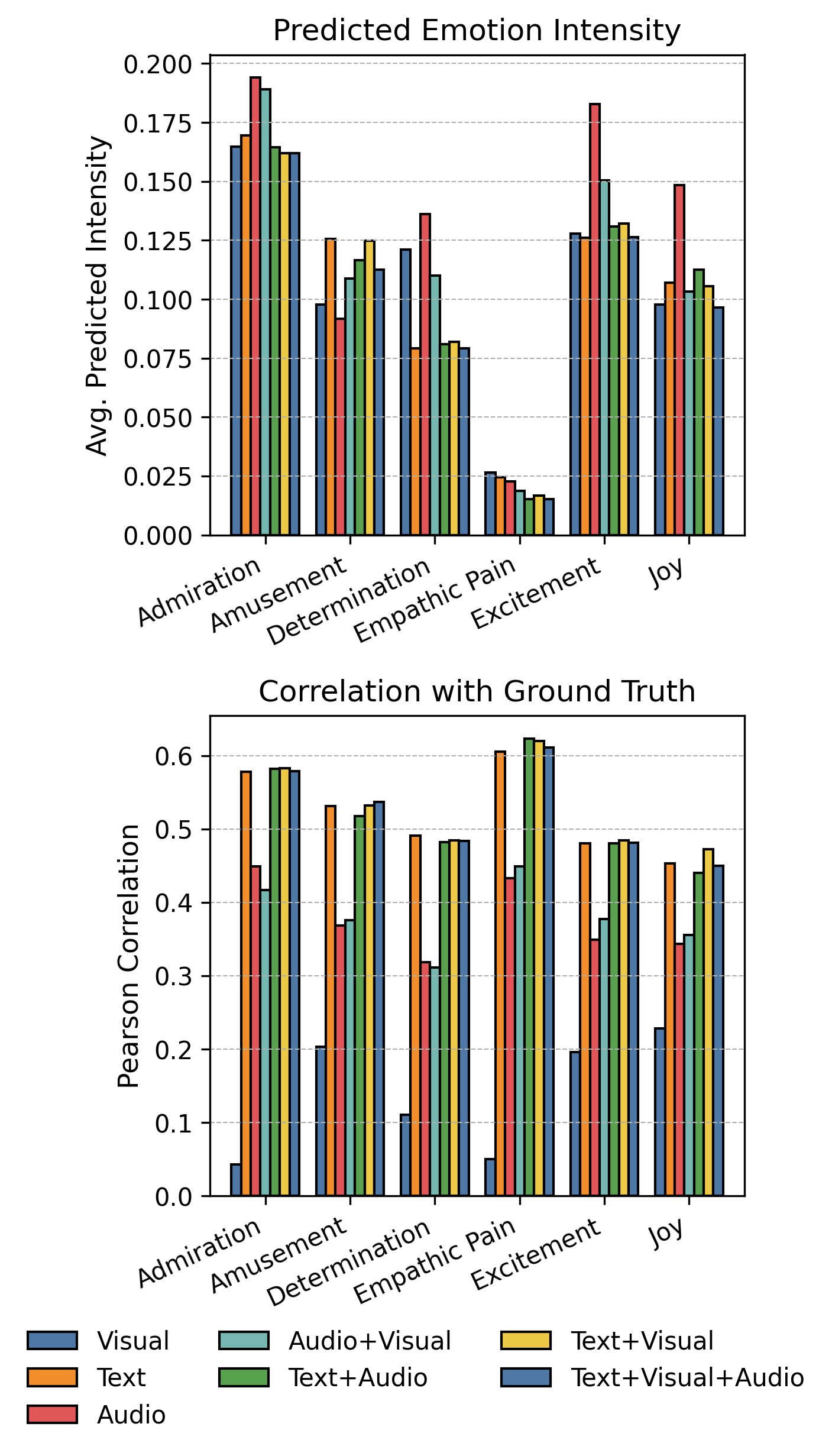}
    \end{center}
      \caption{\textbf{Comparison of the different modalities' performance on the validation split.} Despite being derived from audio, text shows best performance. Vision is the weakest performing modality.}
      \label{fig:emi-modality}
\end{figure}
We additional visualize the predictions as well as the correlation with the ground truth on the validation split in Figure~\ref{fig:emi-modality}. As can be observed in the predicted emotion intensity when using the audio modality alone the model tends to overestimate the emotional intensity in 4 out of 6 emotions. When the visual modality is used exclusively, the determination is significantly higher than for the other modalities and the intensity for amusement is higher for text. Comparing these results with the corresponding correlation; as soon as text is entered as a model input, the results are very homogeneous. 
\begin{table}
    \centering
   \begin{center}
    \begin{tabular}{l|c|c} \hline \hline
       Model  & $\rho_{\text{VAL}}$ & $\rho_{\text{TEST}}$\\ \hline
        Baseline\textsubscript{vision}   & - & .090\\
        Baseline\textsubscript{audio}   & - & .240\\
        \citet{savchenko2025hsemotion} & .446 & .510\\
        \citet{yu2025dual} & .512 & .680\\
        Ours\textsubscript{text}  & .523 & .559\\
        Ours\textsubscript{text+audio}   & .520 & .558\\
        Ours\textsubscript{audio+vision}   & .380 & -\\
        Ours\textsubscript{text+vision}   & \textbf{.529} & .563\\
        Ours\textsubscript{text+audio+vision}   & .522 & .556\\
        Ours\textdagger\textsubscript{text+vision}   & - & \textbf{.706}\\

     \hline \hline
    \end{tabular}
    \end{center}
    \caption{\textbf{Quantitative comparison of our approach on the Hume EMI challenge. The \textdagger~indicates our best model on eval trained on the train and val split.}}
    \label{tab:eval}
\end{table}

\subsubsection{Ablation}
\label{sec:ablationemi}
To determine the best models for predicting emotional mimicry intensity, we evaluated several architectures, as shown in Table~\ref{tab:emiencoder}. Notably, while larger models typically provide better text representations, our results indicate that they perform significantly worse than their base counterparts. This suggests that over-parameterization may lead to overfitting. In contrast, for vision-based models, performance improves with model size, suggesting that larger models provide better representations, but these are not further fine-tuned. For the audio modality, we compared several backbones. Given the strong influence of text on overall performance, we chose WavLM~\cite{chen2022wavlm} and Wav2Vec-BERT~\cite{barrault2023seamless} because of their pre-training goals, which include language-related tasks to improve comprehension. However, our results show that a Wav2Vec 2.0 model~\cite{wagner_2022_6221127} pre-trained on emotional speech gives the best results.
\begin{table}
    \centering
   \begin{center}
    \begin{tabular}{l|cc} \hline \hline
       Model  & $\rho_{\text{VAL}}$ \\ \hline
       \textbf{Text} &\\
        ~~BERT-Base~\cite{devlin2019bert} & .511  \\
        ~~BERT-Large~\cite{devlin2019bert}   & .444 \\
        ~~RoBERTa-Base~\cite{liu2019roberta}  & .497 \\
        ~~GTE-Base~\cite{zhang2024mgte}  & \textbf{.523} \\
        ~~GTE-Large~\cite{zhang2024mgte}   & .390 \\
       \textbf{Vision} &\\
        ~~ViT-Small~\cite{dosovitskiy2020image}   & .086\\
        ~~ViT-Base~\cite{dosovitskiy2020image}   & .132\\
        ~~ViT-Huge~\cite{dosovitskiy2020image}   & \textbf{.138}\\        
       \textbf{Audio} &\\
        ~~W2V2-BERT~\cite{barrault2023seamless}   & .218 \\
        ~~WavLM~\cite{chen2022wavlm}  & .238 \\
        ~~W2V2 Audeering~\cite{wagner_2022_6221127} & \textbf{.376}  \\
     \hline \hline
    \end{tabular}
    \end{center}
    \caption{\textbf{Ablation study for our used encoders.}}
    \label{tab:emiencoder}
\end{table}

\subsubsection{Multi-Task Learning} \label{sec:mtl}
To investigate, why certain modalities, e.g. audio, and combinations thereof, e.g. audio+vision, perform worse than unimodal approaches, cf. Table~\ref{tab:eval} and Figure~\ref{fig:emi-modality}, we tried a multi-task learning (MTL, \cite{kendall2018multi, hallmen2022efficient, hallmen2023phoneme}) approach using the individual tasks' uncertainties for weighing the tasks' losses during training and also as weights for fusing the predictions of each modality. This way we aim to improve the fused prediction overall and also gain insights to possible reasons why including more modalities, thereby more predictive capabilities, counterintuitively performs worse. 

The tasks' uncertainties are learned during training, therefore the architecture in Figure~\ref{fig:bah-approach} has to be slightly altered: instead of fusing all the modalities together before predicting the emotional intensities, each modality is treated as a ``different'' task, i.e. each modality has its own ``fusion module'' making unimodal predictions, which then are fused in an MTL-module using weights to convex-combine the predictions to a final fused prediction.

\begin{table}
    \centering
   \begin{center}
    \begin{tabular}{l|c|ccc} \hline \hline
       Model  & $\rho_{\text{VAL}}$ & $w_\text{Audio}$ & $w_\text{Vision}$ & $w_\text{Text}$ \\ \hline
        Audio+Vision   & .385 & .501 & .499 & - \\
        Audio+Text   & .495 & .498 & - & .502 \\
        Text+Vision   & \textbf{.501} & - & .495 & .505 \\
        Audio+Vision+Text   & .493 & .333 & .331 & .336 \\
     \hline \hline
    \end{tabular}
    \end{center}
    \caption{\textbf{MTL approach investigating each modalities contribution to fused prediction performance.} $w_\text{modality}$ is the contribution weight of a given modality in the prediction process, reflecting its relative influence on the final output.}
    \label{tab:mtl}
\end{table}

The results are in Table~\ref{tab:mtl}. We took the best unimodal models, i.e. GTE-Base, ViT-Huge, and W2V2 Audeering, cf. Table~\ref{tab:bahencoder}, and ran every bi- and trimodal combination thereof. The performance $\rho_{\text{VAL}}$ of MTL achieved lower, but similar levels compared to the base approach without MTL. Consistent with Figure~\ref{fig:emi-modality}, the weight $w$ for text was highest and for audio lowest, i.e. uncertainty for text was lowest and highest for audio. Sadly the models started overfitting before the weights converged, thereby we can only see a tendency here. The MTL approach does work, but because of the architectural changes, i.e. splitting the fusion module and predicting without interdependencies of the modalities, the MTL module cannot leverage the modality interdependent informations.

\subsection{BAH}
Consistent with our EMI challenge experiments, the BAH experiments yielded similar results regarding the validation subset: text is the most informative modality, followed by vision and then audio. As in the EMI experiments, the combination of text and vision also achieved better results on the validation subset than the combination of text, vision, and audio, as demonstrated in Table~\ref{tab:evalbah}.

However, the test subset results for the use of all three modalities - text, vision, and audio - surpass those of text and vision alone. So further investigation is necessary for the BAH task how to fuse the modalities efficiently. For the model trained on the combined validation and train data, another contributing factor to its poor performance is the lack of proper evaluation.

It is noteworthy that a chunk size of 20 seconds yielded the best results for both the text and vision modalities in our experiments. Since the maximum number of image features is limited by hardware constraints and individual frames exhibit little change, 400 equidistant frames are drawn from the 20-second interval. Wav2Vec 2.0 achieves peak performance with a chunk size of 12 seconds in our experiments. 
\begin{table}
    \centering
   \begin{center}
    \begin{tabular}{l|cc} \hline \hline
       Model  & $F1_{\text{VAL}}$ & $F1_{\text{TEST}}$\\ \hline
        Baseline & - & .700 \\
        \citet{savchenko2025hsemotion} & \textbf{.737} & \textbf{.710}\\
        Ours\textsubscript{vision}  & .674 & -\\
        Ours\textsubscript{audio}  & .662 & -\\
        Ours\textsubscript{text}  & .718 & -\\
        Ours\textsubscript{audio+vision}   & .662 & -\\
        Ours\textsubscript{text+audio}   & .696 & -\\
        Ours\textsubscript{text+vision}   & \underline{.725} & .694\\
        Ours\textsubscript{text+audio+vision}   & .720 & \underline{.702}\\
        Ours\textdagger\textsubscript{text+vision}   & - & .693\\     
     \hline \hline
    \end{tabular}
    \end{center}
    \caption{\textbf{Quantitative comparison of our approach on the BAH challenge.
    The \textdagger{} indicates our best model on eval trained on the train and val split.}}
    \label{tab:evalbah}
\end{table}

\subsubsection{Ablation}


Except for the distinction between binary and frame-wise prediction in the BAH challenge, the input data is very similar to that used in the EMI challenge. Therefore, we focused solely on re-evaluating the performance of different text encoders, as shown in Table~\ref{tab:bahencoder}. Once again, GTE~\cite{zhang2024mgte} outperforms BERT~\cite{devlin2019bert} and RoBERTa~\cite{liu2019roberta}. However, in this case, the large transformer model achieves better performance than the base variant, leading us to adopt the large model for the BAH challenge. Since vision and audio provide only marginal performance improvements on the validation split, as shown in Table~\ref{tab:evalbah}, we did not further explore alternative audio or vision encoders for this task.

Table~\ref{tab:textsizebah} illustrates the impact of text chunk size on training performance, i.e., in terms of $F1_{\text{VAL}}$. This analysis was conducted using a text-only training scenario with the GTE-Large model. The results indicate optimal performance at approximately 20 seconds of chunk length, which is why we selected this duration for the text modality in all our experiments during the BAH challenge.

\begin{table}
    \centering
   \begin{center}
    \begin{tabular}{l|cc} \hline \hline
       Model  & $F1_{\text{VAL}}$ \\ \hline
       \textbf{Text} &\\
        ~~BERT-Base~\cite{devlin2019bert} & .701  \\
        ~~BERT-Large~\cite{devlin2019bert}   & .706 \\
        ~~RoBERTa-Base~\cite{liu2019roberta}  & .692 \\
        ~~GTE-Base~\cite{zhang2024mgte}  & .705 \\
        ~~GTE-Large~\cite{zhang2024mgte}   & \textbf{.718} \\
     \hline \hline
    \end{tabular}
    \end{center}
    \caption{\textbf{Ablation study for our used text encoder for the BAH challenge.}}
    \label{tab:bahencoder}
\end{table}

\begin{table}
    \centering
   \begin{center}
    \begin{tabular}{l|cc} \hline \hline
       text chunk size & $F1_{\text{VAL}}$ \\ \hline
        5 seconds & .703  \\
        15 seconds   & .712 \\
        20 seconds  & \textbf{.718} \\
        25 seconds  & .717 \\
        35 seconds   & .716 \\
     \hline \hline
    \end{tabular}
    \end{center}
    \caption{\textbf{Ablation study for our used text chunk size for the BAH challenge, using text as sole modality and GTE-Large as text model.}}
    \label{tab:textsizebah}
\end{table}

\section{Discussion}
Further analysis of these results is necessary, particularly to understand why pre-trained audio models showed limited semantic understanding of the spoken content. During our experiments, we explored alternative audio models, such as Wav2Vec2-BERT~\cite{barrault2023seamless}, which includes an additional language module designed to improve semantic understanding. However, this model failed to effectively capture both semantic and affective dimensions together. We also tested WavLM~\cite{chen2022wavlm}, which, despite being trained on extensive pre-training corpora, similarly struggled to capture the nuanced emotional and semantic information critical to the EMI task. These findings highlight a significant gap in current affective computing research and underscore the need for audio models specifically designed to integrate richer semantic and affective representations.

As we have seen in BAH, on validation the results were consistent with the observations made in EMI on validation, i.e. taking in audio reduces the performance. But when evaluating on test, it switched to the intuitive: more modalities have more possible predictive capabilities, thereby perform equal or better - on test the trimodal approach performed better than the on validation observed bimodal one. This can be seen as a hint, that a trimodal approach can be better on BAH validation, and possibly on EMI validation and test as well.

\section{Future Work}
For future work one could emphasize the tendencies found in the weights learned by MTL, cf. Section~\ref{sec:mtl}, by initializing the task weights not equally, but skewed, e.g. three times more weight for text than other modalities. This could help weights to converge before overfitting of the model takes place, thereby halting training because of deteriorating performance on validation.

An additional approach can be to keep the architecture as-is, run the additional MTL module in parallel instead of replacing the fusion module, and feed the learned weights into the thus slightly extended original fusion module, hopefully improving fusing and not losing the modality interdependent information, that can be leveraged.

\begin{figure}[t]
\begin{center}  \includegraphics[width=\columnwidth]{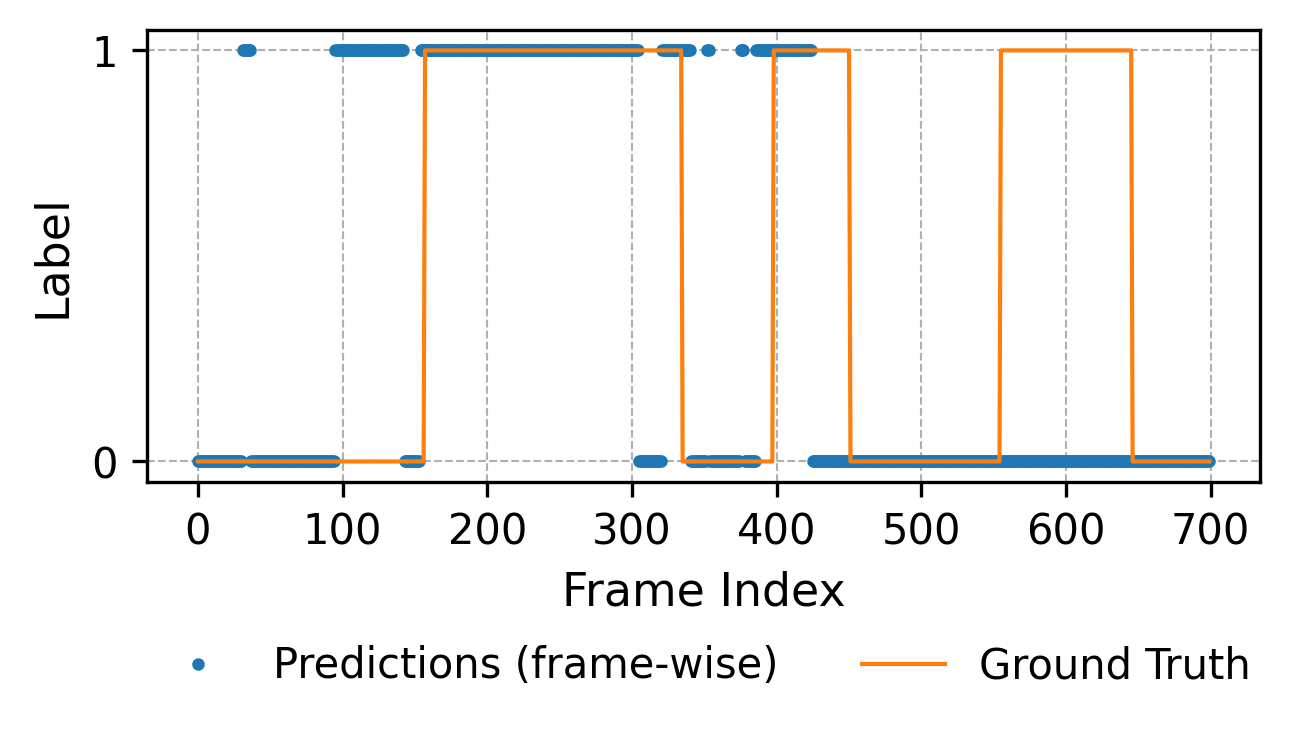}
    \end{center}
        \caption{\textbf{Qualitative example of our predictions with our best model on the BAH task.} While the model detects ambivalence and hesitation, it sometimes shows single-frame positives where there are none. On the other hand, later context is not correctly classified.}
      \label{fig:bah-qualitative}
\end{figure}

Regarding the BAH task, one could explore smoothing techniques to enhance predictions by flattening out high variability in the predictions, based on the assumption, that humans cannot switch from ambivalence/hesitancy every frame, i.e. switch 25 times per second. We have also visualized this behavior of our model in Figure~\ref{fig:bah-qualitative}. Single frames that are false positives, such as in frames 0 to 50, would be removed with further smoothing, while later gaps could be filled.

Since we use sequential context centered around the current frame to predict the class of the current frame, one could try to leverage this approach to a sequence-to-sequence prediction approach. For this recurrent neural networks could either process the whole sequence, or due to hardware constraints, transformers applied on a sliding window.

In our work, we use a simple prediction module to fuse multimodal features. Previous studies~\cite{yu2024efficient, LIU2023679} have demonstrated the significant impact that customized feature fusion can have on performance for affective computing. Future research should investigate more advanced fusion strategies to better exploit complementary information across modalities.

\section{Conclusion}
In this work we have shown that for EMI a combination of textual and visual modality using GTE-Base and ViT-Huge as underlying models works best with a performance on test set of $\rho_{\text{TEST}} = 0.706$, placing first in the EMI challenge.

For BAH, a combination of textual, visual, and auditive modality using GTE-Large, ViT-Huge, and Wav2Vec 2.0 as underlying models works best with a performance on test set of $F1_{\text{TEST}} = 0.702$, placing second in the BAH challenge.

\section*{Acknowledgement}
The authors gratefully acknowledge the computing time granted by the Institute for Distributed Intelligent Systems and provided on the GPU cluster Monacum One at the University of the Bundeswehr Munich.

This work was partially funded by the KodiLL project (FBM2020, Stiftung Innovation in der Hochschullehre).

{
    \small
    \bibliographystyle{ieeenat_fullname}
    \bibliography{main}

\begin{thebibliography}{49}
\providecommand{\natexlab}[1]{#1}
\providecommand{\url}[1]{\texttt{#1}}
\expandafter\ifx\csname urlstyle\endcsname\relax
  \providecommand{\doi}[1]{doi: #1}\else
  \providecommand{\doi}{doi: \begingroup \urlstyle{rm}\Url}\fi

\bibitem[Baevski et~al.(2020)Baevski, Zhou, Mohamed, and Auli]{baevski2020wav2vec}
Alexei Baevski, Yuhao Zhou, Abdelrahman Mohamed, and Michael Auli.
\newblock wav2vec 2.0: A framework for self-supervised learning of speech representations.
\newblock \emph{Advances in neural information processing systems}, 33:\penalty0 12449--12460, 2020.

\bibitem[Barrault et~al.(2023)Barrault, Chung, Meglioli, Dale, Dong, Duppenthaler, Duquenne, Ellis, Elsahar, Haaheim, et~al.]{barrault2023seamless}
Lo{\"\i}c Barrault, Yu-An Chung, Mariano~Coria Meglioli, David Dale, Ning Dong, Mark Duppenthaler, Paul-Ambroise Duquenne, Brian Ellis, Hady Elsahar, Justin Haaheim, et~al.
\newblock Seamless: Multilingual expressive and streaming speech translation.
\newblock \emph{arXiv preprint arXiv:2312.05187}, 2023.

\bibitem[Caron et~al.(2021)Caron, Touvron, Misra, J{\'e}gou, Mairal, Bojanowski, and Joulin]{caron2021emerging}
Mathilde Caron, Hugo Touvron, Ishan Misra, Herv{\'e} J{\'e}gou, Julien Mairal, Piotr Bojanowski, and Armand Joulin.
\newblock Emerging properties in self-supervised vision transformers.
\newblock In \emph{Proceedings of the IEEE/CVF international conference on computer vision}, pages 9650--9660, 2021.

\bibitem[Chen et~al.(2022)Chen, Wang, Chen, Wu, Liu, Chen, Li, Kanda, Yoshioka, Xiao, et~al.]{chen2022wavlm}
Sanyuan Chen, Chengyi Wang, Zhengyang Chen, Yu Wu, Shujie Liu, Zhuo Chen, Jinyu Li, Naoyuki Kanda, Takuya Yoshioka, Xiong Xiao, et~al.
\newblock Wavlm: Large-scale self-supervised pre-training for full stack speech processing.
\newblock \emph{IEEE Journal of Selected Topics in Signal Processing}, 16\penalty0 (6):\penalty0 1505--1518, 2022.

\bibitem[Christ et~al.(2023)Christ, Amiriparian, Baird, Kathan, M{\"u}ller, Klug, Gagne, Tzirakis, Stappen, Me{\ss}ner, et~al.]{christ2023muse}
Lukas Christ, Shahin Amiriparian, Alice Baird, Alexander Kathan, Niklas M{\"u}ller, Steffen Klug, Chris Gagne, Panagiotis Tzirakis, Lukas Stappen, Eva-Maria Me{\ss}ner, et~al.
\newblock The muse 2023 multimodal sentiment analysis challenge: Mimicked emotions, cross-cultural humour, and personalisation.
\newblock In \emph{Proceedings of the 4th on Multimodal Sentiment Analysis Challenge and Workshop: Mimicked Emotions, Humour and Personalisation}, pages 1--10, 2023.

\bibitem[Devlin et~al.(2019)Devlin, Chang, Lee, and Toutanova]{devlin2019bert}
Jacob Devlin, Ming-Wei Chang, Kenton Lee, and Kristina Toutanova.
\newblock Bert: Pre-training of deep bidirectional transformers for language understanding.
\newblock In \emph{Proceedings of the 2019 conference of the North American chapter of the association for computational linguistics: human language technologies, volume 1 (long and short papers)}, pages 4171--4186, 2019.

\bibitem[Dosovitskiy et~al.(2020)Dosovitskiy, Beyer, Kolesnikov, Weissenborn, Zhai, Unterthiner, Dehghani, Minderer, Heigold, Gelly, et~al.]{dosovitskiy2020image}
Alexey Dosovitskiy, Lucas Beyer, Alexander Kolesnikov, Dirk Weissenborn, Xiaohua Zhai, Thomas Unterthiner, Mostafa Dehghani, Matthias Minderer, Georg Heigold, Sylvain Gelly, et~al.
\newblock An image is worth 16x16 words: Transformers for image recognition at scale.
\newblock \emph{arXiv preprint arXiv:2010.11929}, 2020.

\bibitem[Guo et~al.(2016)Guo, Zhang, Hu, He, and Gao]{guo2016ms}
Yandong Guo, Lei Zhang, Yuxiao Hu, Xiaodong He, and Jianfeng Gao.
\newblock Ms-celeb-1m: A dataset and benchmark for large-scale face recognition.
\newblock In \emph{Computer Vision--ECCV 2016: 14th European Conference, Amsterdam, The Netherlands, October 11-14, 2016, Proceedings, Part III 14}, pages 87--102. Springer, 2016.

\bibitem[Hallmen et~al.(2022)Hallmen, Mertes, Schiller, and Andr{\'e}]{hallmen2022efficient}
Tobias Hallmen, Silvan Mertes, Dominik Schiller, and Elisabeth Andr{\'e}.
\newblock An efficient multitask learning architecture for affective vocal burst analysis.
\newblock \emph{arXiv preprint arXiv:2209.13914}, 2022.

\bibitem[Hallmen et~al.(2023)Hallmen, Mertes, Schiller, Lingenfelser, and Andr{\'e}]{hallmen2023phoneme}
Tobias Hallmen, Silvan Mertes, Dominik Schiller, Florian Lingenfelser, and Elisabeth Andr{\'e}.
\newblock Phoneme-based multi-task assessment of affective vocal bursts.
\newblock In \emph{International Conference on Deep Learning Theory and Applications}, pages 209--222. Springer, 2023.

\bibitem[Hallmen et~al.(2024)Hallmen, Deuser, Oswald, and Andr{\'e}]{hallmen2024unimodal}
Tobias Hallmen, Fabian Deuser, Norbert Oswald, and Elisabeth Andr{\'e}.
\newblock Unimodal multi-task fusion for emotional mimicry intensity prediction.
\newblock In \emph{Proceedings of the IEEE/CVF Conference on Computer Vision and Pattern Recognition}, pages 4657--4665, 2024.

\bibitem[He et~al.(2016)He, Zhang, Ren, and Sun]{he2016deep}
Kaiming He, Xiangyu Zhang, Shaoqing Ren, and Jian Sun.
\newblock Deep residual learning for image recognition.
\newblock In \emph{Proceedings of the IEEE conference on computer vision and pattern recognition}, pages 770--778, 2016.

\bibitem[He et~al.(2022)He, Chen, Xie, Li, Doll{\'a}r, and Girshick]{he2022masked}
Kaiming He, Xinlei Chen, Saining Xie, Yanghao Li, Piotr Doll{\'a}r, and Ross Girshick.
\newblock Masked autoencoders are scalable vision learners.
\newblock In \emph{Proceedings of the IEEE/CVF conference on computer vision and pattern recognition}, pages 16000--16009, 2022.

\bibitem[Hershey et~al.(2017)Hershey, Chaudhuri, Ellis, Gemmeke, Jansen, Moore, Plakal, Platt, Saurous, Seybold, et~al.]{hershey2017cnn}
Shawn Hershey, Sourish Chaudhuri, Daniel~PW Ellis, Jort~F Gemmeke, Aren Jansen, R~Channing Moore, Manoj Plakal, Devin Platt, Rif~A Saurous, Bryan Seybold, et~al.
\newblock Cnn architectures for large-scale audio classification.
\newblock In \emph{2017 ieee international conference on acoustics, speech and signal processing (icassp)}, pages 131--135. IEEE, 2017.

\bibitem[Hess and Fischer(2014)]{hess2014emotional}
Ursula Hess and Agneta Fischer.
\newblock Emotional mimicry: Why and when we mimic emotions.
\newblock \emph{Social and personality psychology compass}, 8\penalty0 (2):\penalty0 45--57, 2014.

\bibitem[Hsu et~al.(2021)Hsu, Bolte, Tsai, Lakhotia, Salakhutdinov, and Mohamed]{hsu2021hubert}
Wei-Ning Hsu, Benjamin Bolte, Yao-Hung~Hubert Tsai, Kushal Lakhotia, Ruslan Salakhutdinov, and Abdelrahman Mohamed.
\newblock Hubert: Self-supervised speech representation learning by masked prediction of hidden units.
\newblock \emph{IEEE/ACM transactions on audio, speech, and language processing}, 29:\penalty0 3451--3460, 2021.

\bibitem[Kendall et~al.(2018)Kendall, Gal, and Cipolla]{kendall2018multi}
Alex Kendall, Yarin Gal, and Roberto Cipolla.
\newblock Multi-task learning using uncertainty to weigh losses for scene geometry and semantics.
\newblock In \emph{Proceedings of the IEEE conference on computer vision and pattern recognition}, pages 7482--7491, 2018.

\bibitem[Kollias(2022)]{kollias2022abaw}
Dimitrios Kollias.
\newblock Abaw: Valence-arousal estimation, expression recognition, action unit detection \& multi-task learning challenges.
\newblock In \emph{Proceedings of the IEEE/CVF Conference on Computer Vision and Pattern Recognition}, pages 2328--2336, 2022.

\bibitem[Kollias(2023{\natexlab{a}})]{kollias2023abaw}
Dimitrios Kollias.
\newblock Abaw: learning from synthetic data \& multi-task learning challenges.
\newblock In \emph{European Conference on Computer Vision}, pages 157--172. Springer, 2023{\natexlab{a}}.

\bibitem[Kollias(2023{\natexlab{b}})]{kollias2023multi}
Dimitrios Kollias.
\newblock Multi-label compound expression recognition: C-expr database \& network.
\newblock In \emph{Proceedings of the IEEE/CVF Conference on Computer Vision and Pattern Recognition}, pages 5589--5598, 2023{\natexlab{b}}.

\bibitem[Kollias and Zafeiriou(2019)]{kollias2019expression}
Dimitrios Kollias and Stefanos Zafeiriou.
\newblock Expression, affect, action unit recognition: Aff-wild2, multi-task learning and arcface.
\newblock \emph{arXiv preprint arXiv:1910.04855}, 2019.

\bibitem[Kollias and Zafeiriou(2021{\natexlab{a}})]{kollias2021affect}
Dimitrios Kollias and Stefanos Zafeiriou.
\newblock Affect analysis in-the-wild: Valence-arousal, expressions, action units and a unified framework.
\newblock \emph{arXiv preprint arXiv:2103.15792}, 2021{\natexlab{a}}.

\bibitem[Kollias and Zafeiriou(2021{\natexlab{b}})]{kollias2021analysing}
Dimitrios Kollias and Stefanos Zafeiriou.
\newblock Analysing affective behavior in the second abaw2 competition.
\newblock In \emph{Proceedings of the IEEE/CVF International Conference on Computer Vision}, pages 3652--3660, 2021{\natexlab{b}}.

\bibitem[Kollias et~al.({\natexlab{a}})Kollias, Schulc, Hajiyev, and Zafeiriou]{kollias2020analysing}
D Kollias, A Schulc, E Hajiyev, and S Zafeiriou.
\newblock Analysing affective behavior in the first abaw 2020 competition.
\newblock In \emph{2020 15th IEEE International Conference on Automatic Face and Gesture Recognition (FG 2020)(FG)}, pages 794--800, {\natexlab{a}}.

\bibitem[Kollias et~al.({\natexlab{b}})Kollias, Tzirakis, Cowen, Kotsia, Cogitat, Granger, Pedersoli, Bacon, Baird, Shao, et~al.]{kolliasadvancements}
Dimitrios Kollias, Panagiotis Tzirakis, Alan Cowen, Irene Kotsia, UK Cogitat, Eric Granger, Marco Pedersoli, Simon Bacon, Alice Baird, Chunchang Shao, et~al.
\newblock Advancements in affective and behavior analysis: The 8th abaw workshop and competition.
\newblock {\natexlab{b}}.

\bibitem[Kollias et~al.(2019{\natexlab{a}})Kollias, Sharmanska, and Zafeiriou]{kollias2019face}
Dimitrios Kollias, Viktoriia Sharmanska, and Stefanos Zafeiriou.
\newblock Face behavior a la carte: Expressions, affect and action units in a single network.
\newblock \emph{arXiv preprint arXiv:1910.11111}, 2019{\natexlab{a}}.

\bibitem[Kollias et~al.(2019{\natexlab{b}})Kollias, Tzirakis, Nicolaou, Papaioannou, Zhao, Schuller, Kotsia, and Zafeiriou]{kollias2019deep}
Dimitrios Kollias, Panagiotis Tzirakis, Mihalis~A Nicolaou, Athanasios Papaioannou, Guoying Zhao, Bj{\"o}rn Schuller, Irene Kotsia, and Stefanos Zafeiriou.
\newblock Deep affect prediction in-the-wild: Aff-wild database and challenge, deep architectures, and beyond.
\newblock \emph{International Journal of Computer Vision}, pages 1--23, 2019{\natexlab{b}}.

\bibitem[Kollias et~al.(2023)Kollias, Tzirakis, Baird, Cowen, and Zafeiriou]{kollias2023abaw2}
Dimitrios Kollias, Panagiotis Tzirakis, Alice Baird, Alan Cowen, and Stefanos Zafeiriou.
\newblock Abaw: Valence-arousal estimation, expression recognition, action unit detection \& emotional reaction intensity estimation challenges.
\newblock In \emph{Proceedings of the IEEE/CVF Conference on Computer Vision and Pattern Recognition}, pages 5888--5897, 2023.

\bibitem[Kollias et~al.(2024{\natexlab{a}})Kollias, Sharmanska, and Zafeiriou]{kollias2024distribution}
Dimitrios Kollias, Viktoriia Sharmanska, and Stefanos Zafeiriou.
\newblock Distribution matching for multi-task learning of classification tasks: a large-scale study on faces \& beyond.
\newblock In \emph{Proceedings of the AAAI Conference on Artificial Intelligence}, pages 2813--2821, 2024{\natexlab{a}}.

\bibitem[Kollias et~al.(2024{\natexlab{b}})Kollias, Tzirakis, Cowen, Zafeiriou, Shao, and Hu]{kollias20246th}
Dimitrios Kollias, Panagiotis Tzirakis, Alan Cowen, Stefanos Zafeiriou, Chunchang Shao, and Guanyu Hu.
\newblock The 6th affective behavior analysis in-the-wild (abaw) competition.
\newblock \emph{arXiv preprint arXiv:2402.19344}, 2024{\natexlab{b}}.

\bibitem[Kollias et~al.(2024{\natexlab{c}})Kollias, Zafeiriou, Kotsia, Dhall, Ghosh, Shao, and Hu]{kollias20247th}
Dimitrios Kollias, Stefanos Zafeiriou, Irene Kotsia, Abhinav Dhall, Shreya Ghosh, Chunchang Shao, and Guanyu Hu.
\newblock 7th abaw competition: Multi-task learning and compound expression recognition.
\newblock \emph{arXiv preprint arXiv:2407.03835}, 2024{\natexlab{c}}.

\bibitem[Kollias et~al.(2025)Kollias, Tzirakis, Cowen, Zafeiriou, Kotsia, Granger, Pedersoli, Bacon, Baird, Gagne, Shao, Hu, Belharbi, and Aslam]{Kollias2025}
Dimitrios Kollias, Panagiotis Tzirakis, Alan~S. Cowen, Stefanos Zafeiriou, Irene Kotsia, Eric Granger, Marco Pedersoli, Simon~L. Bacon, Alice Baird, Chris Gagne, Chunchang Shao, Guanyu Hu, Soufiane Belharbi, and Muhammad~Haseeb Aslam.
\newblock {Advancements in Affective and Behavior Analysis: The 8th ABAW Workshop and Competition}.
\newblock 2025.

\bibitem[Liu et~al.(2023)Liu, Gao, Li, Fu, and Ding]{LIU2023679}
Shuai Liu, Peng Gao, Yating Li, Weina Fu, and Weiping Ding.
\newblock Multi-modal fusion network with complementarity and importance for emotion recognition.
\newblock \emph{Information Sciences}, 619:\penalty0 679--694, 2023.

\bibitem[Liu et~al.(2019)Liu, Ott, Goyal, Du, Joshi, Chen, Levy, Lewis, Zettlemoyer, and Stoyanov]{liu2019roberta}
Yinhan Liu, Myle Ott, Naman Goyal, Jingfei Du, Mandar Joshi, Danqi Chen, Omer Levy, Mike Lewis, Luke Zettlemoyer, and Veselin Stoyanov.
\newblock Roberta: A robustly optimized bert pretraining approach.
\newblock \emph{arXiv preprint arXiv:1907.11692}, 2019.

\bibitem[Lotfian and Busso(2019)]{Lotfian_2019_3}
R. Lotfian and C. Busso.
\newblock Building naturalistic emotionally balanced speech corpus by retrieving emotional speech from existing podcast recordings.
\newblock \emph{IEEE Transactions on Affective Computing}, 10\penalty0 (4):\penalty0 471--483, 2019.

\bibitem[Radford et~al.(2023)Radford, Kim, Xu, Brockman, McLeavey, and Sutskever]{radford2023robust}
Alec Radford, Jong~Wook Kim, Tao Xu, Greg Brockman, Christine McLeavey, and Ilya Sutskever.
\newblock Robust speech recognition via large-scale weak supervision.
\newblock In \emph{International conference on machine learning}, pages 28492--28518. PMLR, 2023.

\bibitem[Richet et~al.(2024)Richet, Belharbi, Aslam, Schadt, Gonz{\'a}lez-Gonz{\'a}lez, Cortal, Koerich, Pedersoli, Finkel, Bacon, et~al.]{richet2024textualized}
Nicolas Richet, Soufiane Belharbi, Haseeb Aslam, Meike~Emilie Schadt, Manuela Gonz{\'a}lez-Gonz{\'a}lez, Gustave Cortal, Alessandro~Lameiras Koerich, Marco Pedersoli, Alain Finkel, Simon Bacon, et~al.
\newblock Textualized and feature-based models for compound multimodal emotion recognition in the wild.
\newblock \emph{arXiv preprint arXiv:2407.12927}, 2024.

\bibitem[Savchenko(2023)]{savchenko2023facial}
Andrey Savchenko.
\newblock Facial expression recognition with adaptive frame rate based on multiple testing correction.
\newblock In \emph{Proceedings of the 40th International Conference on Machine Learning (ICML)}, pages 30119--30129. PMLR, 2023.

\bibitem[Savchenko(2024)]{savchenko2024hsemotion}
Andrey~V Savchenko.
\newblock Hsemotion team at the 6th abaw competition: Facial expressions, valence-arousal and emotion intensity prediction.
\newblock \emph{arXiv preprint arXiv:2403.11590}, 2024.

\bibitem[Savchenko(2025)]{savchenko2025hsemotion}
Andrey~V Savchenko.
\newblock Hsemotion team at abaw-8 competition: Audiovisual ambivalence/hesitancy, emotional mimicry intensity and facial expression recognition.
\newblock \emph{arXiv preprint arXiv:2503.10399}, 2025.

\bibitem[Sutskever et~al.(2014)Sutskever, Vinyals, and Le]{sutskever2014sequence}
Ilya Sutskever, Oriol Vinyals, and Quoc~V Le.
\newblock Sequence to sequence learning with neural networks.
\newblock \emph{Advances in neural information processing systems}, 27, 2014.

\bibitem[Wagner et~al.(2022)Wagner, Triantafyllopoulos, Wierstorf, Schmitt, Burkhardt, Eyben, and Schuller]{wagner_2022_6221127}
Johannes Wagner, Andreas Triantafyllopoulos, Hagen Wierstorf, Maximilian Schmitt, Felix Burkhardt, Florian Eyben, and Björn~W. Schuller.
\newblock {Model for Dimensional Speech Emotion Recognition based on Wav2vec 2.0}, 2022.

\bibitem[Wu et~al.(2020)Wu, Xu, Dai, Wan, Zhang, Yan, Tomizuka, Gonzalez, Keutzer, and Vajda]{wu2020visual}
Bichen Wu, Chenfeng Xu, Xiaoliang Dai, Alvin Wan, Peizhao Zhang, Zhicheng Yan, Masayoshi Tomizuka, Joseph Gonzalez, Kurt Keutzer, and Peter Vajda.
\newblock Visual transformers: Token-based image representation and processing for computer vision, 2020.

\bibitem[Yu et~al.(2024)Yu, Zhu, Zhu, Cai, Zhao, Zhang, Xie, Wei, Liu, and Liang]{yu2024efficient}
Jun Yu, Wangyuan Zhu, Jichao Zhu, Zhongpeng Cai, Gongpeng Zhao, Zerui Zhang, Guochen Xie, Zhihong Wei, Qingsong Liu, and Jiaen Liang.
\newblock Efficient feature extraction and late fusion strategy for audiovisual emotional mimicry intensity estimation.
\newblock In \emph{Proceedings of the IEEE/CVF Conference on Computer Vision and Pattern Recognition}, pages 4866--4872, 2024.

\bibitem[Yu et~al.(2025)Yu, Zhu, Chi, Zhang, Zheng, Wang, and Lu]{yu2025dual}
Jun Yu, Lingsi Zhu, Yanjun Chi, Yunxiang Zhang, Yang Zheng, Yongqi Wang, and Xilong Lu.
\newblock Dual-stage cross-modal network with dynamic feature fusion for emotional mimicry intensity estimation.
\newblock \emph{arXiv preprint arXiv:2503.10603}, 2025.

\bibitem[Zafeiriou et~al.(2017)Zafeiriou, Kollias, Nicolaou, Papaioannou, Zhao, and Kotsia]{zafeiriou2017aff}
Stefanos Zafeiriou, Dimitrios Kollias, Mihalis~A Nicolaou, Athanasios Papaioannou, Guoying Zhao, and Irene Kotsia.
\newblock Aff-wild: Valence and arousal ‘in-the-wild’challenge.
\newblock In \emph{Computer Vision and Pattern Recognition Workshops (CVPRW), 2017 IEEE Conference on}, pages 1980--1987. IEEE, 2017.

\bibitem[Zhang et~al.(2022)Zhang, An, Ding, and Guan]{zhang2022continuous}
Su Zhang, Ruyi An, Yi Ding, and Cuntai Guan.
\newblock Continuous emotion recognition using visual-audio-linguistic information: A technical report for abaw3.
\newblock In \emph{Proceedings of the IEEE/CVF Conference on Computer Vision and Pattern Recognition}, pages 2376--2381, 2022.

\bibitem[Zhang et~al.(2024{\natexlab{a}})Zhang, Qiu, Liu, Li, Du, Guo, and Yu]{zhang2024affective}
Wei Zhang, Feng Qiu, Chen Liu, Lincheng Li, Heming Du, Tiancheng Guo, and Xin Yu.
\newblock Affective behaviour analysis via integrating multi-modal knowledge.
\newblock \emph{arXiv preprint arXiv:2403.10825}, 2024{\natexlab{a}}.

\bibitem[Zhang et~al.(2024{\natexlab{b}})Zhang, Zhang, Long, Xie, Dai, Tang, Lin, Yang, Xie, Huang, et~al.]{zhang2024mgte}
Xin Zhang, Yanzhao Zhang, Dingkun Long, Wen Xie, Ziqi Dai, Jialong Tang, Huan Lin, Baosong Yang, Pengjun Xie, Fei Huang, et~al.
\newblock mgte: Generalized long-context text representation and reranking models for multilingual text retrieval.
\newblock \emph{arXiv preprint arXiv:2407.19669}, 2024{\natexlab{b}}.

\end{thebibliography}
}


\end{document}